\pdfoutput=1
\documentclass[11pt]{article}

\usepackage{acl}

\usepackage{times}
\usepackage{latexsym}
\usepackage{hyperref}
\usepackage{xurl}

\usepackage[T1]{fontenc}

\usepackage[utf8]{inputenc}

\usepackage{microtype}

\usepackage{graphicx}
\usepackage{booktabs}
\usepackage{tabularx}
\usepackage{multirow}
\usepackage{subcaption}
\usepackage{makecell}
\usepackage{listings}
\usepackage[finalizecache=false,frozencache=true,cachedir=minted-cache]{minted}

\title{Few-Shot Semantic Parsing with Language Models Trained on Code}

\author{Richard Shin \\
  Microsoft Semantic Machines \\
  \texttt{richard.shin@microsoft.com} \\\And
  Benjamin Van Durme \\
  Microsoft Semantic Machines \\
  \texttt{ben.vandurme@microsoft.com} \\}

\begin{document}
\maketitle
\begin{abstract}
Large language models can perform semantic parsing with little training data, when prompted with in-context examples.
It has been shown that this can be improved by formulating the problem as paraphrasing into \emph{canonical utterances}, which casts the underlying meaning representation into a controlled natural language-like representation.
Intuitively, such models can more easily output canonical utterances as they are closer to the natural language used for pre-training.
Recently, models also pre-trained on code, like OpenAI Codex, have risen in prominence.
For semantic parsing tasks where we map natural language into code, such models may prove more adept at it.
In this paper, we test this hypothesis and find that Codex performs better on such tasks than equivalent GPT-3 models.
We evaluate on Overnight and SMCalFlow and find that unlike GPT-3, Codex performs similarly when targeting meaning representations directly,
perhaps because meaning representations are structured similar to code
in these datasets.

\end{abstract}

\section{Introduction}
Semantic parsing is the task of mapping natural language to a target meaning representation.
Many approaches have been explored by the community, including a recent focus on the use of large autoregressive language models (LMs).
Such pretrained LMs 
can achieve 
surprising levels of accuracy with relatively small numbers of examples.
Further gains have 
come from
constraining a decoder to only consider syntactically valid outputs.

Historically, \emph{language} models have been constructed using a large collection of \emph{natural} language.
And yet, the term ``language'' clearly applies to non-natural languages as well.
Very large models 
have been
trained on mixed corpora, explicitly curated to include code (programming language) as well as natural language.
Examples include GPT-J \citep{gpt-j}, MT-NLG \citep{mt-nlg}, and Gopher \citep{rae2021gopher-arxiv}, with OpenAI Codex \citep{2107-03374}, PaLM-Coder \cite{palm}, and \citet{2108-07732} particularly focused on code.

We revisit few-shot semantic parsing experiments from \citet{shin-etal-2021-constrained},
which used GPT-3 with constrained decoding into a controlled sublanguage of English (canonical utterances) 
then translated the canonical utterance output into the meaning representation using a synchronous context-free grammar (SCFG).
In this work, we perform similar experiments on the Overnight \citep{wang-etal-2015-building} and SMCalFlow \citep{andreas-etal-2020-task} datasets,\footnote{Both are in English and available under CC BY-SA 4.0.}
but using OpenAI Codex instead. 
As Codex has been trained on code,
including natural language comments that explain its intent,
we hypothesize that Codex will be more adept at semantic parsing for meaning representations resembling code.

In this work, we find that:

 \begin{list}{$\bullet$}
  { \setlength{\itemsep}{0pt}
     \setlength{\parsep}{2pt}
     \setlength{\topsep}{2pt}
     \setlength{\partopsep}{0pt}
     \setlength{\leftmargin}{1.1em}
     \setlength{\labelwidth}{0.7em}
     \setlength{\labelsep}{0.4em}}
\item Codex substantially narrows the gap in accuracy between predicting meaning representations directly versus canonical utterances, thus obviating the need to define canonical utterances. %
We observe this even though the meaning representations use bespoke languages, rather than ones like Python which frequently appeared in the training data.
\item Surprisingly, Codex also generates canonical utterances more accurately than GPT-3, even though those look more like English than code.
\item Even with Codex, constrained decoding with a CFG and a non-greedy search procedure are still valuable in providing improved accuracy.
\item \emph{Speculative constrained decoding}, an adaptation of \citet[Appendix F]{poesia2022synchromesh}, gives comparable accuracy as beam search but with greater efficiency, on the language model APIs provided by OpenAI.
\end{list}

\begin{table*}[t]
\small
\centering
\begin{tabularx}{\textwidth}{lp{2.5cm}p{2.6cm}X}
\toprule
Dataset & Natural language & Canonical utterance & Meaning representation \\
\midrule
SMCalFlow & Schedule Hide and Seek in the mall for Saturday night & create event called "Hide and Seek" starting next Saturday night at "mall" & \texttt{\scriptsize (Yield :output (CreateCommitEventWrapper :event (CreatePreflightEventWrapper :constraint (Constraint[Event] :subject (?= \#(String "Hide and Seek")) :start (DateTimeConstraint :constraint (Night) :date (NextDOW :dow \#(DayOfWeek "SATURDAY"))) :location (?= \#(LocationKeyphrase "mall"))))))} \\
Overnight Cal. & which meeting has the earliest end time & meeting that has the smallest end time & \texttt{\scriptsize (call listValue (call superlative (call getProperty (call singleton en.meeting) (string !type)) (string min) (call ensureNumericProperty (string end\_time))))} \\
\bottomrule
\end{tabularx}
\caption{Examples from the Overnight Calendar and SMCalFlow datasets.}
\label{tab:data-examples}
\end{table*}

\section{Preliminaries}
\subsection{Constrained language model parsing}
\label{sec:clamp}
In semantic parsing, our goal is to convert an utterance $u$ into the meaning representation $m$.
We use the same approach as \citet{shin-etal-2021-constrained}:
(1) priming the underlying language model with dynamically created prompts,
(2) constrained decoder,
and (3) optionally using a canonical utterance $c$ as the target output instead of $m$.

Since GPT-3 and Codex can perform \emph{in-context few-shot learning} \citep{brown2020language},
we retrieve 20 $(u_i, m_i)$ pairs most similar\footnote{
    We use GPT-3 itself for this, following \citet{shin-etal-2021-constrained}.
    The similarity function is identical for all our experiments,
    regardless of whether we use GPT-3 or Codex for decoding.
}
to $u$ from the training set,
then translate $m_i$ into $c_i$ if using canonical utterances,
to form the prompt $p$ which looks like:
\begin{lstlisting}
Let's translate what a human user says into what a computer might say.

Human: when is the standup $\leftarrow u_1$
Computer: start time of "standup" $\leftarrow c_1$
Human: what date is the standup $\leftarrow u_2$
Computer: date of "standup" $\leftarrow c_2$
$[...]$
Human: how long is the daily standup $\leftarrow u$
Computer:
\end{lstlisting}
where $italics$ are annotations for exposition in this paper, and not included verbatim in the prompt.

We then generate a completion for $p$ using the language model, which we will take as the predicted value of canonical utterance $c$ or meaning representation $m$, depending on our choice for (3).
To ensure that the generated completion is well-formed,
we assume the existence of a function $\texttt{nextTokens}(s) = \{w_i\} \subseteq \mathcal{V} \cup \{\texttt{EOS}\}$. For a given prefix $s$ of a canonical utterance or meaning representation, this function returns the set of subsequent tokens that we can can append to $s$ such that it remains a prefix of a well-formed $c$ or $m$. It also indicates whether $s$ is already a complete, well-formed $c$ or $m$ by including \texttt{EOS} in the result; if $\texttt{nextTokens}(s) = \{\texttt{EOS}\}$, then $s$ is a valid canonical utterance or meaning representation with no possible extensions.

As an example, \texttt{nextTokens}(``\texttt{start time}'') would contain \texttt{of},
but not \texttt{EOS} or \texttt{in}.
We use \texttt{nextTokens} to filter candidates from the language model such that
it only generates grammatical outputs;
if we build the completion by appending what $\texttt{nextTokens}$ advises, we are guaranteed to obtain a grammatically conformant output.
We implement $\texttt{nextTokens}$ using a trie and a CFG for Overnight and SMCalFlow, respectively.

\subsection{OpenAI language models}
\label{sec:openai-lm}
OpenAI operates a service offering GPT-3 \citep{brown2020language} through a networked API.
The API includes multiple variants of GPT-3, named Ada, Babbage, Curie, and Davinci,
with the model size increasing in that order.
Two Codex \citep{2107-03374} models, which had code from GitHub included in their training data,
are also offered. They are named Cushman Codex and Davinci Codex.\footnote{
We used the models available in late 2021; OpenAI may change them from time to time.}

The primary use case for the API is generating completions from a prefix,
by sequentially sampling from $p(w_{n} | w_1 w_2 \cdots w_{n-1})$
until some limit is reached.
The API provides for specifying a softmax temperature to modify this distribution,
for example enabling greedy argmax sampling with a temperature of 0.0.
The API also allows for directly querying $p(w_{n} | w_1 w_2 \cdots w_{n-1})$,
but only returns probabilities for up to 100 most likely tokens;
we use this capability for constrained beam search.

\subsection{Experimental setup}
\label{sec:exp-setup}
We used two of the datasets from \citet{shin-etal-2021-constrained} for our experiments. %
We build on their released code and use the same subsets of the training data.
We briefly describe some of the details below.

\paragraph{Overnight.}
This dataset from \citet{wang-etal-2015-building} contains 13,682 examples across eight different domains,
curated to exhibit a variety of linguistic phenomena and semantic structures.
We used 200 randomly-sampled training examples for each domain, and evaluate on the domains separately.
For evaluation, we use denotational accuracy, based on comparing the execution results of the predicted and reference programs.

\paragraph{SMCalFlow.}
Introduced in \citet{andreas-etal-2020-task},
this task-oriented dialogue dataset consists of conversations about calendars, weather, places, and people.
Each utterance $u$ is annotated with dataflow programs $m$ containing function composition, complex constraints,
and references to computations from previous turns.
Of the 133,821 $(u_i, m_i)$ pairs in training, we use a stratified sample of 300 for our experiments,
following \citet{shin-etal-2021-constrained}.
For evaluation, we use syntactical match between the predicted and reference programs, which requires them to be structurally identical but allows differences of spacing and named arguments in function calls.

\paragraph{Test set sampling for certain experiments.}
As usage of GPT-3 and Codex requires significant resources,
we conduct our initial experiments on smaller subsets of the evaluation sets.
For Overnight, we used 100 uniformly sampled examples from test set for the calendar domain.
For SMCalFlow, we used 200 uniformly sampled examples from the validation set.

We used the subsets for the experiments described in Sections~\ref{sec:gpt3-vs-codex} to \ref{sec:speculative-decoding}.
In the final experiments of Section~\ref{sec:exp-final}, we use the full test set for Overnight and the full validation set for SMCalFlow.

\section{Experiments}
\subsection{Comparing GPT-3 and Codex}
\label{sec:gpt3-vs-codex}

\begin{table}[ht]
\small
\centering
\begin{tabular}{lrr}
\toprule
& \multicolumn{2}{c}{Accuracy} \\
Model & Overnight Cal. & SMCalFlow \\
\midrule
Davinci & 0.81 & 0.340 \\
Curie & 0.66 & 0.260 \\
Davinci Codex & 0.86 & 0.355 \\
Cushman Codex & 0.87 & 0.320 \\

\bottomrule
\end{tabular}
\caption{Comparing various OpenAI models using constrained
decoding to generate canonical utterances, with beam search
having beam size 5. These results are on 100 sampled test
examples.  The larger Davinci models do better, the Codex models
show better performance.}
\label{tab:codex-vs-gpt3}
\end{table}

Table~\ref{tab:codex-vs-gpt3} summarizes our initial comparison of the GPT-3 and Codex models when applied to semantic parsing.
Davinci Codex performs better than Davinci on both Overnight Calendar and SMCalFlow when using identical settings.
More interestingly, Cushman Codex, which is one step down from Davinci Codex, performs substantially better than Curie, which is one step down from Davinci.
These results support our hypothesis that language models trained on code can perform better at semantic parsing.

\subsection{Targeting canonical utterances versus meaning representations}
\label{sec:canon-vs-mr}

\begin{table}[h]
\small
\centering

\begin{subtable}[h]{\columnwidth}
\centering
\begin{tabular}{lrrr}
\toprule
& \multicolumn{3}{c}{Accuracy} \\
Model & Canonical & Meaning & $C - M$ \\
\midrule
Davinci & 0.81 & 0.68 & 0.13 \\
Davinci Codex & 0.86 & 0.86 & 0.00 \\

\bottomrule
\end{tabular}
\caption{Overnight Calendar}
\end{subtable}

\begin{subtable}[h]{\columnwidth}
\centering
\begin{tabular}{lrrr}
\toprule
& \multicolumn{3}{c}{Accuracy} \\
Model & Canonical & Meaning & $C - M$ \\
\midrule
Davinci & 0.340 & 0.245 & 0.095 \\
Davinci Codex & 0.355 & 0.345 & 0.010 \\

\bottomrule
\end{tabular}
\caption{SMCalFlow}
\end{subtable}

\caption{Differences in accuracy when using canonical utterances
versus directly using meaning representations.  Davinci Codex
performs better on canonical utterances, but the gap is much
smaller than with Davinci.  Results using constrained decoding
with beam size 5.}
\label{tab:canon-vs-mr}
\end{table}

\citet{shin-etal-2021-constrained} demonstrated that as language models have (conventionally) been trained to generate natural language,
we would benefit by formulating semantic parsing as paraphrasing into a controlled sublanguage of English.
In Table~\ref{tab:canon-vs-mr}, we investigate whether that still holds true when using Codex.
We observe that when using GPT-3 (Davinci), targeting meaning representations can result in more than a 25\% relative drop in accuracy.
In contrast, Davinci Codex exhibits no or a very small drop in accuracy when targeting meaning representations.

Notably, the meaning representations used for Overnight and SMCalFlow are in Lisp-like languages,
rather than in programming languages common on GitHub.
Our experiments indicate that Codex can nevertheless pick up on the semantics with only a few examples in the prompt.

Having canonical utterances as the target output still performs better than meaning representations.
This holds true even though our evaluation procedure first translates canonical utterances back into meaning representations, which is a lossy procedure for SMCalFlow as described in \citep{shin-etal-2021-constrained}. 
However, designing a suitable system of canonical utterances is a non-trivial effort. The smaller performance gap we observe with Codex changes the cost/benefit calculations on authoring SCFGs.

\subsection{Value of constraints and beam search}
\label{sec:exp-constraints-bs}
As mentioned in Section~\ref{sec:openai-lm},
the primary capability of OpenAI's API is generating completions from a prefix using sequential sampling.
Their documentation\footnote{\url{https://beta.openai.com/docs/guides/completion/working-with-code}} suggests using it that way
to generate code from comments, a  similar task to semantic parsing. 
Nevertheless, we see in Table~\ref{tab:constrained-vs-unconstrained}
that the use of constraints and beam search lead to benefits in accuracy.
Even with constrained decoding, greedy argmax sampling (equivalent to a beam size of 1) performs worse than using beam search.

\begin{table}[h]
\small
\centering
\begin{tabular}{lrrr}
\toprule
& & \multicolumn{2}{c}{Accuracy} \\
Decoding & Beam & Overnight Cal. & SMCalFlow \\
\midrule
Constrained & 5 & 0.86 & 0.345 \\
Constrained & 1 & 0.75 & 0.300 \\
Unconstrained & 5 & 0.80 & 0.315 \\
Unconstrained & 1 & 0.73 & 0.280 \\

\bottomrule
\end{tabular}
\caption{Results comparing constrained with unconstrained decoding and multiple beam sizes, when generating meaning representations.
Even when using Davinci Codex, trained specifically on code, constrained decoding and beam search lead to higher accuracy.}
\label{tab:constrained-vs-unconstrained}
\end{table}

\begin{table*}[ht]
\small
\centering
\begin{tabular}{rrrrrrrrrr}
\toprule
& & \multicolumn{4}{c}{Overnight Calendar} & \multicolumn{4}{c}{SMCalFlow} \\
& & \multicolumn{2}{c}{Accuracy} & \multicolumn{2}{c}{Items/second} & \multicolumn{2}{c}{Accuracy} & \multicolumn{2}{c}{Items/second}\\
Width & Temperature & Canonical & Meaning & Canonical & Meaning & Canonical & Meaning & Canonical & Meaning \\
\midrule
1 & 0.0 & 0.86 & 0.76 & 0.520 & 0.246 & 0.300 & 0.320 & 0.193 & 0.184 \\
1 & BS & 0.84 & 0.75 & 0.237 & 0.059 & 0.305 & 0.300 & 0.116 & 0.040 \\
5 & 0.5 & 0.87 & 0.80 & 0.380 & 0.155 & 0.335 & 0.315 & 0.076 & 0.140 \\
5 & 1.0 & 0.87 & 0.85 & 0.260 & 0.145 & 0.325 & 0.330 & 0.076 & 0.034 \\
5 & BS & 0.86 & 0.86 & 0.133 & 0.030 & 0.355 & 0.345 & 0.065 & 0.008 \\
10 & 0.5 & 0.87 & 0.86 & 0.355 & 0.150 & 0.345 & 0.345 & 0.038 & 0.085 \\
10 & 1.0 & 0.87 & 0.85 & 0.193 & 0.068 & 0.370 & 0.335 & 0.028 & 0.014 \\

\bottomrule
\end{tabular}

\caption{Comparing various settings on speculative constrained decoding with beam search.
``BS'' indicates use of beam search.
Speculative constrained decoding gets similar accuracy as beam search, but at higher speed.}
\label{tab:spec-settings}
\end{table*}

\subsection{Speculative constrained decoding}
\label{sec:speculative-decoding}
While constrained decoding and beam search improve accuracy,
they are slow to perform with OpenAI's API.
Extending a partial hypothesis requires one network round-trip per token.
The API lacks state and so each request includes the prompt and all previously generated tokens.
In the worst case, the statelessness implies decoding will take $O(n^3)$ complexity rather than the typical $O(n^2)$ of transformers
due to needing to re-encode the prefix each time.
Even if the hidden states for previous tokens were cached, their retrieval and  transfer to GPUs or other accelerators takes overhead.

As such, we adapt a method from Synchromesh \citep[Appendix F]{poesia2022synchromesh}
to obtain the benefits of beam search and constrained decoding with greater efficiency.
We extend Synchromesh's approach with a \emph{width} parameter $W$, which functions similar to the beam size.
We call it \emph{speculative constrained decoding}.

To expand a partial hypothesis in the search,
we query the API to create $W$ completions with softmax temperature $T$.\footnote{The softmax function with temperature $T$ computes $\frac{\exp(x_i/T)}{\sum_{j=1}^{|V|} \exp(x_j/T)}$, to compute probabilities for each of the $|V|$ tokens in the vocabulary.
As $T$ approaches 0, the output becomes 1 for the largest value of $x_i$ and $0$ for all others, effectively computing the argmax.
}
The API samples from the model, without reference to any grammars,
until \texttt{EOS} is sampled or a length limit is reached.
Using the \texttt{nextTokens} function, we check each of the $W$ completions left-to-right until we encounter an invalid token,
and truncate there so that we only have valid tokens; we return the truncated completions as new hypotheses.
If no completion contains any valid tokens, then we query the API for the $W$ best tokens and return those as new hypotheses.
As done in beam search, we start with a single empty hypothesis, and keep the $W$ best expansions at each step.
We stop after 16 steps if $W$ complete hypotheses were not generated by then.
More details are in Appendix~\ref{sec:spec-decoding-algo}.

Table~\ref{tab:spec-settings} shows the results from trying various values for $W$ and $T$, along with beam search for $W = 1$ and $W = 5$.
When $W = 1$ and $T = 0$, which is equivalent to Synchromesh's approach,
we obtain very similar results to constrained greedy decoding (beam size 1).
However, speculative constrained decoding is substantially faster.

In order to obtain results comparable to beam search with beam size 5, we require $W = 5$ or $10$.
In comparison, Synchromesh only supports $W = 1$.
We see notable speedups compared to beam search, but typically obtain comparable accuracy.

We also observe that we can generate generate canonical utterances more quickly than meaning representations, as the canonical utterances are shorter. However, these timing results do not include the time required to convert canonical utterances into meaning representations.

\subsection{Putting everything together}
\label{sec:exp-final}

\begin{table}[h]
\centering
\small
\begin{tabular}{lrr}
\toprule
& \multicolumn{2}{c}{Accuracy} \\
Model & Overnight Avg. & SMCalFlow \\
\midrule
\makecell[l]{\citet{shin-etal-2021-constrained}, \\ \ \ Constrained Canonical} & 0.765 & 0.32 \\
\makecell[l]{\citet{shin-etal-2021-constrained}, \\ \ \ Constrained Meaning} & 0.657* & 0.25* \\
Ours, Canonical & 0.785 & 0.342 \\
Ours, Meaning & 0.750 & 0.330 \\

\bottomrule
\end{tabular}
\caption{Comparison to \citet{shin-etal-2021-constrained}.
Results are on the entire test set for Overnight and the entire dev set for SMCalFlow.
For Overnight, we took a simple average of the accuracy for each of the 8 domains.
Results marked with * are on subsampled evaluation sets.
We used speculative constrained decoding with a width of 10 and a temperature of 0.5.}
\label{tab:final}
\end{table}

As explained in Section~\ref{sec:exp-setup},
earlier results in this article are based on smaller subsets of the evaluation sets due to resource limitations.
In Table~\ref{tab:final}, we evaluate on the full evaluation sets using lessons learned from our previous experiments.
We achieve better accuracies than when \citet{shin-etal-2021-constrained} used GPT-3.
We re-confirm Section~\ref{sec:canon-vs-mr} that Codex performs nearly as well at meaning representations as canonical utterances.

\section{Related Work}
\newcite{chen-etal-2020-low} observed that 
for low-resource semantic parsing, fine-tuning a pretrained sequence-to-sequence model
improved over the use of a pretrained encoder only. \newcite{scholak-etal-2021-picard}, \newcite{wu-etal-2021-paraphrasing}, and \newcite{shin-etal-2021-constrained} each proposed the use of constrained decoding for semantic parsing with LMs. The latter two works argued that language models were best used to parse language into controlled natural language, rather than directly to a code-like representation. Here we consider whether that conclusion changes based on new LMs that are trained with code.

\newcite{pasupat-etal-2021-controllable} proposed a retrieval-augmented solution to semantic parsing, which relates to the dynamic prompt selection of \newcite{shin-etal-2021-constrained}, and which we followed here without alteration. Future work may consider the impact of more advanced prompt selection techniques.

\section{Conclusion}
We investigate the use of OpenAI Codex, a large language model trained on code, for few-shot semantic parsing.
We find that it performs better than GPT-3 for our tasks.
While constrained decoding and a non-greedy decoding procedure still non-trivially improve accuracy,
mapping to canonical natural language is no longer as important with Codex, thereby lightening the burden on developing few shot semantic parsers based on large LMs.

\section*{Ethical Considerations}
Our work heavily relies on OpenAI's GPT-3 and Codex models, which are large language models trained on big datasets.
Such language models may reflect biases present in their training data \citep{brown2020language,10.1145/3442188.3445922}.
However, our use of constrained decoding largely mitigates the risks from such bias as we only allow the model to generate outputs allowed by a small grammar.
Furthermore, the outputs are interpreted by machines rather than directly shown to humans.
The potential for harm may increase when the grammars used in constrained decoding allow for a wider variety of outputs (such as including unconstrained free-text fields),
and if semantic parsing is used for particularly sensitive domains.

\section*{Acknowledgements}
We would like to thank Adam Pauls, Jason Eisner, Matt Gardner, and other colleagues at Microsoft Semantic Machines, who provided helpful feedback and engaged in stimulating discussions which have greatly improved the paper.

\bibliography{anthology,custom}
\bibliographystyle{acl_natbib}

\appendix

\section{Measuring performance of beam search and speculative constrained decoding}
For measuring the items/second of beam search and speculative constrained decoding in Table~\ref{tab:spec-settings} and Table~\ref{tab:spec-settings-full}, we used the first 10 items of the evaluation sets.
As we only had access to shared instances of GPT-3 and Codex, we were unable to guarantee lack of interference from other users.
While the numbers are not precise, we believe they are generally indicative of the expected performance of the two methods.

\section{Prompt for Codex when using meaning representations}
Instead of the prompt in Section~\ref{sec:clamp}, we used the prompt depicted below:

\begin{lstlisting}[frame=single]
;;; Translate questions into Lisp expressions

; [utterance from training example]
[meaning representation from example]
; [utterance from training example]
[meaning representation from example]
$[...]$
; [test utterance]
\end{lstlisting}

The text in square brackets are for exposition and not included verbatim in the prompt.

\section{Supplementary results}

\begin{table*}[ht]
\centering
\begin{tabular}{lllrrr}
\toprule
& & & & \multicolumn{2}{c}{Accuracy} \\
Model & Output & Decoding & Beam size & Overnight Cal. & SMCalFlow \\
\midrule
Davinci & Canonical & Constrained & 5 & 0.81 & 0.340 \\
Davinci & Canonical & Constrained & 1 & 0.76 & 0.290 \\
Davinci & Canonical & Unconstrained & 5 & 0.72 & 0.295 \\
Davinci & Canonical & Unconstrained & 1 & 0.72 & 0.255 \\
Davinci & Meaning & Constrained & 5 & 0.68 & 0.245 \\
Davinci & Meaning & Constrained & 1 & 0.62 & 0.210 \\
Davinci & Meaning & Unconstrained & 5 & 0.53 & 0.230 \\
Davinci & Meaning & Unconstrained & 1 & 0.48 & 0.190 \\
Curie & Canonical & Constrained & 5 & 0.66 & 0.260 \\
Curie & Canonical & Constrained & 1 & 0.58 & 0.210 \\
Curie & Canonical & Unconstrained & 5 & 0.50 & 0.225 \\
Curie & Canonical & Unconstrained & 1 & 0.47 & 0.210 \\
Curie & Meaning & Constrained & 5 & 0.44 & 0.200 \\
Curie & Meaning & Constrained & 1 & 0.39 & 0.165 \\
Curie & Meaning & Unconstrained & 5 & 0.38 & 0.185 \\
Curie & Meaning & Unconstrained & 1 & 0.31 & 0.160 \\
Davinci Codex & Canonical & Constrained & 5 & 0.86 & 0.355 \\
Davinci Codex & Canonical & Constrained & 1 & 0.84 & 0.305 \\
Davinci Codex & Canonical & Unconstrained & 5 & 0.79 & 0.310 \\
Davinci Codex & Canonical & Unconstrained & 1 & 0.77 & 0.295 \\
Davinci Codex & Meaning & Constrained & 5 & 0.86 & 0.345 \\
Davinci Codex & Meaning & Constrained & 1 & 0.75 & 0.300 \\
Davinci Codex & Meaning & Unconstrained & 5 & 0.80 & 0.315 \\
Davinci Codex & Meaning & Unconstrained & 1 & 0.73 & 0.280 \\
Cushman Codex & Canonical & Constrained & 5 & 0.87 & 0.320 \\
Cushman Codex & Canonical & Constrained & 1 & 0.80 & 0.290 \\
Cushman Codex & Canonical & Unconstrained & 5 & 0.83 & 0.300 \\
Cushman Codex & Canonical & Unconstrained & 1 & 0.77 & 0.285 \\
Cushman Codex & Meaning & Constrained & 5 & 0.80 & 0.340 \\
Cushman Codex & Meaning & Constrained & 1 & 0.73 & 0.280 \\
Cushman Codex & Meaning & Unconstrained & 5 & 0.72 & 0.305 \\
Cushman Codex & Meaning & Unconstrained & 1 & 0.70 & 0.250 \\

\bottomrule
\end{tabular}
\caption{All results on Overnight Calendar and SMCalFlow using beam search.}
\label{tab:mega}
\end{table*}

\begin{table*}[h]
\small
\centering
\begin{tabular}{rrrrrrrrrr}
\toprule
& & \multicolumn{4}{c}{Overnight Calendar} & \multicolumn{4}{c}{SMCalFlow} \\
& & \multicolumn{2}{c}{Accuracy} & \multicolumn{2}{c}{Items/second} & \multicolumn{2}{c}{Accuracy} & \multicolumn{2}{c}{Items/second}\\
Width & Temperature & Canonical & Meaning & Canonical & Meaning & Canonical & Meaning & Canonical & Meaning \\
\midrule
1 & 0.0 & 0.86 & 0.76 & 0.520 & 0.246 & 0.300 & 0.320 & 0.193 & 0.184 \\
1 & BS & 0.840 & 0.750 & 0.237 & 0.059 & 0.305 & 0.300 & 0.116 & 0.040 \\
5 & 0.25 & 0.86 & 0.79 & 0.553 & 0.208 & 0.330 & 0.325 & 0.071 & 0.050 \\
5 & 0.5 & 0.87 & 0.80 & 0.380 & 0.155 & 0.335 & 0.315 & 0.076 & 0.140 \\
5 & 0.75 & 0.86 & 0.84 & 0.344 & 0.129 & 0.320 & 0.340 & 0.076 & 0.081 \\
5 & 1.0 & 0.87 & 0.85 & 0.260 & 0.145 & 0.325 & 0.330 & 0.076 & 0.034 \\
5 & BS & 0.860 & 0.860 & 0.133 & 0.030 & 0.355 & 0.345 & 0.065 & 0.008 \\
10 & 0.25 & 0.88 & 0.81 & 0.537 & 0.213 & 0.345 & 0.310 & 0.020 & 0.040 \\
10 & 0.5 & 0.87 & 0.86 & 0.355 & 0.150 & 0.345 & 0.345 & 0.038 & 0.085 \\
10 & 0.75 & 0.87 & 0.82 & 0.266 & 0.103 & 0.350 & 0.355 & 0.039 & 0.034 \\
10 & 1.0 & 0.87 & 0.85 & 0.193 & 0.068 & 0.370 & 0.335 & 0.028 & 0.014 \\

\bottomrule
\end{tabular}

\caption{Comparing various settings on speculative decoding with beam search.
``BS'' for temperature indicates use of beam search. This table is an expanded version of 
Table \ref{tab:spec-settings}}
\label{tab:spec-settings-full}
\end{table*}

Table~\ref{tab:mega} contains all results from using beam search, used to construct Tables \ref{tab:codex-vs-gpt3}, \ref{tab:canon-vs-mr}, and \ref{tab:constrained-vs-unconstrained}.
Table~\ref{tab:spec-settings-full} is a version of Table~\ref{tab:spec-settings} with more rows.

\section{Speculative constrained decoding algorithm}
\label{sec:spec-decoding-algo}
To further expand on the description in Section~\ref{sec:speculative-decoding},
we express the speculative constrained decoding method in Python-like pseudocode in Listing~\ref{listing:spec-decoding-algo}.

\begin{listing*}[t]
\begin{minted}[fontsize=\footnotesize,mathescape]{python}
# Parameters:
# - W = width of the search
# - T = softmax temepature
# - MAX_STEPS = How many times we invoke the model. We set this to 16.
#
# Helper functions:
# - nextTokens: as defined in Section 2.1
# - model_completions: ask the model to generate completions with the given
#   prefix. Returns a list of token sequences sampled after the prefix.
# - length_normalized_logprob: compute the log probability of a token sequence,
#   where longer sequences receive a bonus.
# - is_finished: check if a token sequence is finished according to the grammar.
#
# `search` is invoked with tokens for the prompt $p$ for a given example.

def expand(tokens):
    samples = model_completions(tokens, temperature=T, num_completions=W)

    results = []
    for sample in samples:
        valid_prefix = tokens
        for token in sample:
            if token not in nextTokens(prefix):
                break
            valid_prefix += [token]
        if valid_prefix == tokens:
            # No tokens in the completion were grammatically valid.
            # Back off to regular constrained decoding to advance by one token,
            # and append to results
            ...
        else:
            results += [valid_prefix]
    return results


def search(prompt):
    # We start with one hypothesis containing tokens from the prompt.
    beam = [prompt] 
    finished = []

    for _ in range(MAX_STEPS):
        candidates = []
        for state in beam:
            candidates += expand(state)
        candidates.sort(key=length_normalized_logprob, reverse=True)

        new_beam = []
        for cand in candidates:
            if is_finished(cand):
                finished.append(cand)
            else:
                new_beam.append(cand)
            if len(finished) + len(new_beam) == W:
                break
        
        if len(new_beam) == 0:
            break
        else:
            beam = new_beam

    return finished
\end{minted}
\caption{Pseudocode for speculative constrained decoding}
\label{listing:spec-decoding-algo}
\end{listing*}

\end{document}